\documentclass[preprint,sigconf,noacm]{acmart}
\setcopyright{none}
\renewcommand\footnotetextcopyrightpermission[1]{}

\AtBeginDocument{%
  \providecommand\BibTeX{{%
    \normalfont B\kern-0.5em{\scshape i\kern-0.25em b}\kern-0.8em\TeX}}}

\acmConference{SUD'21: Supporting and Understanding of Conversational Dialogues}{March 12, 2021}{Jerusalem, Israel}
\acmBooktitle{}
\acmPrice{}
\acmISBN{}

\usepackage{booktabs}
\usepackage{multirow}

\begin{document}

\title[Explaining Outcomes of Multi-Party Dialogues using Causal Learning]{Explaining Outcomes of Multi-Party Dialogues\\using Causal Learning}\titlenote{Presented at SUD'21: Supporting and Understanding of Conversational Dialogues, March 12, 2021, Jerusalem, Israel. SUD'21 workshop was held at WSDM'21: The 14th ACM International Conference on Web Search and Data Mining}
\author{Priyanka Sinha}
\email{priyanka.sinha.iitg@gmail.com}
\orcid{0000-0002-7971-4449}
\affiliation{
\institution{Tata Consultancy Services Limited}
\institution{Indian Institute of Technology Kharagpur}
\city{Kolkata}
\country{India}
}
\author{Pabitra Mitra}
\email{pabitra@gmail.com}
\orcid{0000-0002-1908-9813}
\affiliation{
\institution{Indian Institute of Technology Kharagpur}
\city{Kharagpur}
\country{India}
}
\author{Antonio Anastasio Bruto da Costa}
\email{antonio.cse.iitkgp@gmail.com}
\orcid{0000-0002-4590-0665}
\affiliation{%
  \institution{Indian Institute of Technology Kharagpur}
  \city{Kharagpur}
  \country{India}
}
\author{Nikolaos Kekatos}
\email{nikos.kekatos@gmail.com}
\orcid{0000-0002-9421-8566}
\affiliation{%
  \institution{Aristotle University of Thessaloniki}
  \city{Thessaloniki}
  \country{Greece}
 }

\renewcommand{\shortauthors}{Priyanka, Pabitra, Antonio and Nikolaos}

\begin{abstract}
  Multi-party dialogues are common in enterprise social media on
technical as well as non-technical topics. The outcome of a conversation
may be positive or negative. It is important to analyze why a dialogue
ends with a particular sentiment from the point of view of conflict
analysis as well as future collaboration design. We propose an
explainable time series mining algorithm for such analysis. A dialogue
is represented as an attributed time series of occurrences of
keywords, {\tt EMPATH} categories, and inferred sentiments at various points in its progress. 
A special decision tree, with decision metrics that take into account temporal relationships between dialogue events, is used for predicting the cause of the outcome sentiment. Interpretable rules mined from the classifier are used to explain the prediction. Experimental results
are presented for the enterprise social media posts in a large
company.

\end{abstract}



\begin{teaserfigure}
  \includegraphics[width=\textwidth]{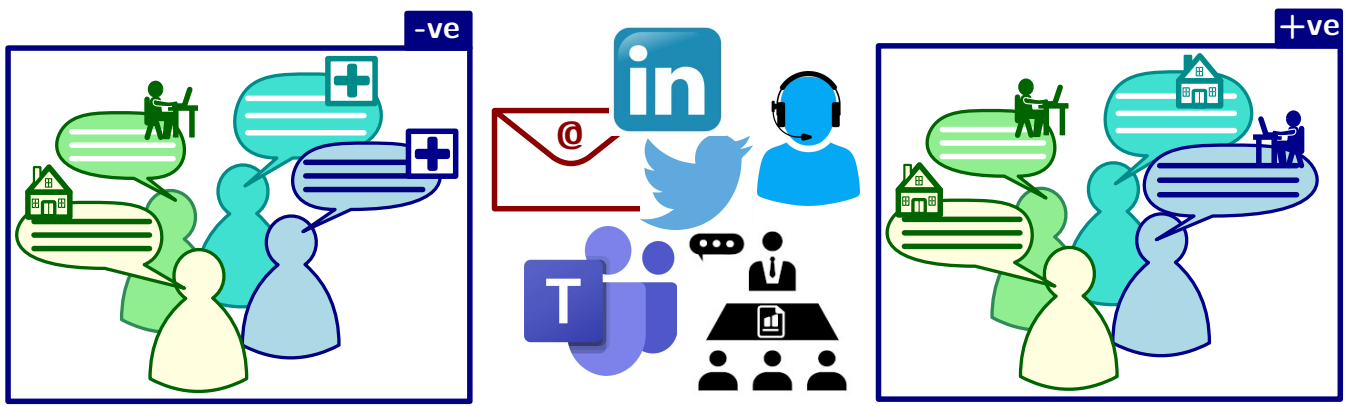}
  \caption{Multi-party dialogue outcomes are explainable from psycholinguistic features; dialogues triggered by medical emergencies lead to a negative outcome (-ve), while dialogues regarding family and work matters lead to a positive outcome (+ve).}
  \Description{Multiple people in enterprise social media setting having a dialogue with different psychological states resulting in varied outcome.}
  \label{fig:teaser}
\end{teaserfigure}

\maketitle


\section{Introduction}

Linguistic interaction between two or more people is a primary form of communication. Most research in language focuses on the two-party case, e.g. communication between two people, a person and a dialogue system, or a pair of agents~\cite{traum2003issues}. Some work has also been done on multi-party dialogues~\cite{ishizaki1998exploring,Shi_Huang_2019,cscw2006}.

In general, in a dialogue involving two or more parties, the participants express various sentiments and evoke multiple psychological processes in the parties involved~\cite{pickering2006dance}. Two important characteristics of dialogues are the \emph{timing} of response and the natural need for \emph{alignment} of authors' psychological states~\cite{pickering2020understanding}. The large body of conversations on the web, like social media, calls, and meeting transcripts, are unobtrusive traces of dialogues that provide secondary data for psychological experiments. This data can be used to automate the understanding of the psychology of dialogue.




Feeling emotions is an essential characteristic that differentiates humans from machines. Analyzing and understanding these human emotions are particularly important and have motivated the emergence of several research fields centered on the analysis of sentiments in communications. 

Sentiment analysis in customer service dialogue can be beneficial in various applications, such as service satisfaction analysis and intelligent agent applications~\cite{devillers2006real}. The authors of~\cite{wang2020sentiment}
 work on sentiment classification in customer service dialogue, which aims to assign a proper sentiment label to each response in a customer service dialogue.

More specifically, in call centers, it is undesirable for the companies to have customers that leave the call dissatisfied. Most call transcripts are stored and can be studied to better analyze and identify what caused the customer to be displeased. The additional options given to customers to provide numerical ratings could be used alongside sentiment analysis to obtain a clearer understanding of the situation, what went wrong, and provide directions for service improvements in the future. 


Similarly, transcripts or stored text/audio traces from multi-party dialogues can be found in business and board room meetings. This information can be very useful for the company and its stakeholders to better understand its employees' behaviors and feelings and be better prepared in the future.


On Enterprise Social Media (ESM), it is always desirable for the discussion to converge to a positive sentiment even if there are arguments. This matter is very significant in discussions around a company policy. A dissatisfied or angry conclusion may cause difficulties in implementing the policy. It is observed that people with certain personalities may not get along well with each other. They may either generally be disagreeable people or work better with people who have different personalities. For instance, a highly conscientious person may demonstrate anger towards a dishonest person in a dialogue between these two. ESM, through their posts, comments, likes, and other attributes provide an unobtrusive sensor to the cognitive state of the person. These comments and posts can be seen as dialogue traces. In fact, the timestamped sequence of a post and its comments may be considered as a multi-party dialogue. In turn, it can be represented as a discrete-time series of attributes of each post or comment. These attributes may include the users/authors' sentiment polarity and intensity, popularity, and psycholinguistic processes. The same may also be viewed from the perspective of a single user, and the timeline of posting activity of a user could be represented as a discrete-time series of similar attributes. In this case, the timestamps directly mark the timing of response of the participants in such a dialogue. Text mining techniques are able to extract the cognitive state of the authors. Therefore this could be used as a substitute for live experiments by psychologists \cite{pickering2020understanding}. 

These multi-party dialogues may be viewed as temporal sequences of messages by participants. We are motivated to mine explainable  dialogue rules automatically. These rules must causally explain the outcomes of interactions of parties in a dialogue in a psychologically meaningful manner that may contribute to our understanding of dialogues themselves. For psychologists, some of these may further be verified in their experimental settings. For people designing collaboration tools, these rules may help them evaluate their designs. For team managers in an enterprise, these may provide actionable insights into group behavior.

In this paper, we address the research question as to what sequences of psychological states and sentiments in multi-party dialogues result in a positive or negative outcome. We further understand the cause of dialogues that received appreciation in terms of "thank" or apologies in terms of "sorry" and yet do not have a positive outcome. This leads us to address what author psychological states and sentiment values contribute to either scenario. The novelty of our work is that we are able to incorporate timing in explaining causal effects in dialogues, unlike previous work, e.g.~\cite{pechsiri2007mining}, where they generate rules that have no timing or sequence information.

The organization of the remainder of the paper is as follows. In Section~\ref{sec:algo}, we propose a comprehensive methodology to obtain explanations for the outcomes from dialogue texts using causal learning. In Section~\ref{sec:exper}, we apply the proposed methodology in a large dataset taken from ESM. We provide concluding remarks in Section~\ref{sec:concl}.

\section{Proposed Methodology} \label{sec:algo}

\begin{figure*}[t!]
    \centering
    \includegraphics[width=\textwidth]{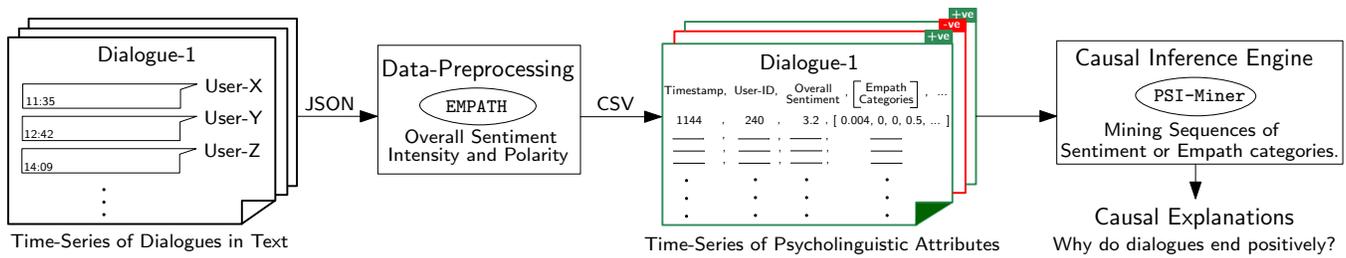}
    \caption{Pipeline for explaining outcomes of multi-party dialogues.}
    \label{fig:methodology}
\end{figure*}

In this section, we discuss our methodology for mining causal explanations for a dialogue outcome. The methodology is inspired by studies in psycholinguistics and symbolic methods for causal learning. The methodology pipeline is outlined in Figure~\ref{fig:methodology}. 

A dataset containing dialogues, like time series, is the input to the pipeline. A dialogue is a collection of tuples. Each tuple contains i) a timestamp, ii) text in natural language, and iii) the ID of the text's author. 

There are two core steps to our method. 
\begin{description}
    \item[Step 1] In the first step, the psycholinguistic attributes of the author that indicate the cognitive state of the author's mind, are textually mined from their messages as part of the dialogues. For a dialogue, this step aims to mine psychologically relevant attributes from each text entry. For instance, for a text entry, some attributes of interest could be whether the author uses words to indicate that they are thinking about a medical emergency, domestic work, office work, or is expressing fear or talking about their family. Such attributes are known as psycholinguistic attributes. For a text entry, a numerical measure is assigned to each attribute to indicate either its absence or a positive value indicative of the author's intensity as mined from the text. A dialogue containing a sequence of messages by multiple authors, i.e., unstructured text, is transformed into a dialogue with quantitative evaluations for each text entry. The outcome of this step is a file containing timestamped comma-separated-values for the attributes. 
    \item[Step 2] The second step uses the time series of psycholinguistic attributes to learn explanations for dialogue outcomes. Although explanations could be learned for various attributes, in this article we focus our analysis on whether a dialogue concludes on a positive or negative note. We use a causal analysis framework that learns cause-effect statements from time series. The framework takes the sequence as well as the timing of events in the time series into account while constructing the cause. The outcome of this step is statements in \emph{symbolic temporal logic} that are both human and machine interpretable.
\end{description}

Each of these steps are further elaborated in Sections~\ref{sec:psycho-linguistic-features} and~\ref{sec:causal-learning}.
\subsection{Psycho-linguistic Features}\label{sec:psycho-linguistic-features}

Dialogue text can be either written, e.g. in social media, or spoken speech that is transcribed into text, e.g. meeting transcripts and customer service call transcripts. In both forms, dialogue text is prone to noise like misspellings, grammatical errors, not fully formed sentences, emoticons, short text, domain specific acronyms and other such features. The presence of such noise makes the use of traditional linguistic processing approaches perform poorly. This mainly happens as traditional approaches assume well-formed long text.

Sentiment polarity and intensity of a post by an author reveals both their stance in the dialogue as well as their opinion towards the participants. There are several sentiment lexicons that are available, many of which have sparse coverage in the noisy text that we work within this paper. To mine sentiments expressed by the author from noisy text, we use a fast Naive Bayes based sentiment analyzer in the Python package {\tt TextBlob} \cite{textblob}. Despite existing deep neural network based sentiment analyzers, we find {\tt TextBlob} to be fast and efficient for our purpose, as we have observed from \cite{www2014} that Naive Bayes based sentiment analyzer would perform well in noisy text. We use pretrained models without the need for re-training on our dataset. The values of sentiment intensity and polarity would explain our inference mined from dialogue traces when we obtain a meaningful sequence such as a consequent sequence of high positive sentiment comments leading to a positive conclusion despite a few low negative sentiments expressed within. It would also help us uncover authors who despite regular negative sentiment do not affect the overall tone and concluding sentiment of a dialogue. This would be useful in customer service situations to identify intrinsically dissatisfied customers and in enterprise contexts, naysayers who would perhaps benefit from a change of team or role.

Studies have shown that the choice of words has a clear psychological basis and reveals the authors' state of mind. In Linguistic Inquiry and Word Count (LIWC) \cite{liwc}, it is shown that 
the use of certain words indicates the presence or absence of a category of psychological processes of the author. The results are validated via experiments based on essays by participants in a situated survey. Noise makes it harder to easily infer word usage from the LIWC dictionary. The work on EMPATH \cite{empath} extends \cite{liwc} by using a deep neural network based word embedding technique to generate a model for identifying similar categories in such noisy text. In the pretrained model of EMPATH that we employ in this paper, it is possible to mine the intensity of about 200 human validated categories that are help, office, dance, money, wedding, domestic work, sleep, medical emergency, cold, hate, cheerfulness, aggression, occupation, envy, anticipation, family, vacation, crime, attractive, masculine, prison, health, pride, dispute, nervousness, government, weakness, horror, swearing terms, leisure, suffering, royalty, wealthy, tourism, furniture, school, magic, beach, journalism, morning, banking, social media, exercise, night, kill, blue collar job, art, ridicule, play, computer, college, optimism, stealing, real estate, home, divine, sexual, fear, irritability, superhero, business, driving, pet, childish, cooking, exasperation, religion, hipster, internet, surprise, reading, worship, leader, independence, movement, body, noise, eating, medieval, zest, confusion, water, sports, death, healing, legend, heroic, celebration, restaurant, violence, programming, dominant hierarchical, military, neglect, swimming, exotic, love, hiking, communication, hearing, order, sympathy, hygiene, weather, anonymity, trust, ancient, deception, fabric, air travel, fight, dominant personality, music, vehicle, politeness, toy, farming, meeting, war, speaking, listen, urban, shopping, disgust, fire, tool, phone, gain, sound, injury, sailing, rage, science, work, appearance, valuable, warmth, youth, sadness, fun, emotional, joy, affection, traveling, fashion, ugliness, lust, shame, torment, economics, anger, politics, ship, clothing, car, strength, technology, breaking, shape and size, power, white collar job, animal, party, terrorism, smell, disappointment, poor, plant, pain, beauty, timidity, philosophy, negotiate, negative emotion, cleaning, messaging, competing, law, friends, payment, achievement, alcohol, liquid, feminine, weapon, children, monster, ocean, giving, contentment, writing, rural, positive emotion, musical.
Using all these categories can better explain our inferences mined from dialogue traces. An example of a meaningful sequence that we can obtain is that high aggression comments in a meeting context  conclude on a negative note. It also helps us uncover uncommon sequences and adds to our understanding of author psychology in a dialogue setting. 

In this work, we focus on specific words that are relevant to the enterprise context that have known behavioral interpretation. Words like `thank' and `sorry' are used by an author to express their gratitude towards another. In our ESM dataset, these two words are found to be used frequently. In an enterprise context, these words of gratitude are usually used to show appreciation towards someone. Appreciation is important from a human resource perspective and is correlated with improved job satisfaction. The word `thank' is used to appreciate someone. When used in a comment, the author is thanking the other author and is a mark of appreciation. Similarly, the word `sorry' is used to apologize to someone. In an enterprise, this is rarely used, as it shows humility and the author may fear looking weak especially from an assessment perspective. 

ESM datasets also provide relevant nonlinguistic attributes such as popularity, likes, shares, follows that have behavioral significance. These additional attributes can have a big impact on the dialogue's outcome. Let us consider an example of a dialogue ending up positively despite the existence of multiple existing negative comments. An inferred explanation could be that there exists one very influential positive comment which attracted more "likes" and overcame all the intermediate negative comments. 

\subsection{Mining Explanations for Dialogue Outcomes using Causal Learning}\label{sec:causal-learning}
The timing and order of certain interactions in a dialogue determine its outcome. For instance, one would say that a negative response of \textit{User-X} to a comment by \textit{User-Y} occurring within the same day has an impact on the negativity of the outcome, while if the negative response is on the next day, it doesn't have a role in determining the outcome. It is important that such interaction timings are factored into the learning process. However, across dialogues, the timing of such interactions that appear to cause a particular kind of outcome, may be different. In the phrase "\textit{interaction event B happens within an interval of time after interaction event A}", we say that $B$ \textit{non-deterministically} occurs within an interval of time after the occurrence $A$. Learning the timing between interactions must take this non-determinism into account.

The tool {\tt PSIMiner}~\cite{jair_CostaD21} enables the easy learning of causal relationships of this form from a chosen set of features. The unique advantage of using {\tt PSIMiner} is its ability to learn an explanation of an outcome as a timed sequence of features. It uses specially developed decision metrics to predict a cause for a given target. The prediction is an interpretable temporal logic statement. This is especially useful in the context of dialogues, which include interactions over time, and where the order of interactions, their psycho-linguistic features, and their separation in time, all contribute to an outcome sentiment. 

{\tt PSIMiner} allows us to provide a flexible explanation template, as a sequence of events over time.  The form of an explanation template is as follows:
\begin{center}
{\tt
 $\sqcup_n$ \#\#[0:k]  $\sqcup_{n-1}$ \#\#[0:k]...\#\#[0:k] $\sqcup_1$ \#\#[0:k] $\sqcup_0$ |-> E }
\end{center}
The template takes the form "\textit{cause} $\rightarrow$ \textit{effect}". Each $\sqcup$ is called a \textit{bucket}. A bucket is a placeholder and may contain zero or more features. $n$ and $k$ are meta-parameters of {\tt PSIMiner}. There are $n+1$ buckets in the template above. Each adjacent pair of buckets are separated by a time interval of up to $k$ time units. The sequence of buckets forms the cause. The effect $E$ is a placeholder for the outcome. The explanation template reads in English as follows, "\textit{\textbf{If} $\sqcup_n$ happens, and then within zero to k time units $\sqcup_{n-1}$ happens, and then ..., and then $\sqcup_0$ happens, \textbf{then} $E$ happens}". {\tt PSIMiner} predicts which features go into which buckets, and may also leave some buckets empty. If buckets are left empty, then the time intervals on both sides of an empty bucket merge into a larger time interval. From the dataset, it will refine the time separations between buckets.  

In the context of a dialogue, the buckets in the template correspond to distinctive interaction features, while the time intervals represent the timed order of the interactions. These interactions once taken together determine the outcome. For example consider the explanation, $\langle$\textit{User-X,progress}$\rangle$ {\tt \#\#[30:167]} $\langle$\textit{gratitude}$\rangle$,  for a positive dialogue outcome. The explanation says that \textit{User-X} expressed that they had progressed in work and 30 to 167 time units later gratitude was expressed. In this example, the user that expressed the gratitude doesn't factor into the explanation because it may not be important who expresses gratitude. This can be because dialogues exist where different users express gratitude to \textit{User-X}.

The inputs to {\tt PSIMiner} include a set of time series (with data formatted into comma-separated-values (CSV)), a set of features expressed as predicates over the variables in the time-series, and meta-parameters for learning. The meta-parameters determine the explanation template and include values for $n$ (the number of time-intervals) and $k$ (the initial limit on each time-interval). The reader may recall that the number of buckets in the template is one more than the number of time-intervals. An additional meta-parameter is used to place a bound on the depth of the decision tree built by {\tt PSIMiner}. This in turn implicitly places a bound on the maximum number of features used to form an explanation. This bound on the number of features  controls how feature complex an explanation can become and protects against data over-fitting. 

In addition to providing values for $n$, $k$ and the depth of the decision tree, a set of predicates, representing knowledge, may also be provided. The outcome, which is the target of the explanation, is one such predicate. 

The output of {\tt PSIMiner} is a list of rules in the format specified by the template. Each rule of the form "\textit{cause~$\rightarrow$~effect}" is associated with a \textit{correlation} metric~\cite{jair_CostaD21}. This metric measures how much of the effect is attributed to the cause. For instance, for the rule $S \rightarrow E$, a correlation of $40\%$ indicates that the cause $S$ contributes to $40\%$ of the observations of effect $E$ in the data. Rules with higher correlation are preferred.  

\section{Experimental Results}\label{sec:exper}

We demonstrate the effectiveness of our proposed methodology on an ESM dataset, which is described in Section \ref{sec:dataset}. Section~\ref{sec:dialoguePreProcessing} describes how we pre-process the dataset to transform the social media dialogues into trace files used as inputs to the causal inference engine. The learned explanation rules for dialogue outcomes are presented in Section~\ref{sec:results} where we also qualitatively discuss the human interpretation of the rules and their practical soundness.

\subsection{Dataset} \label{sec:dataset}

We have collected social media content from an enterprises' internal custom ESM platform which uses a REST API and the OAuth protocol. For privacy reasons, we are unable to disclose the name of the custom made platform for the enterprise, used internally. This enterprise consists of over 450,000 white collar employees spread across 130 countries with over 34 percent of them being women. All employees participate and interact in the ESM. However, the volume of engagement with the platform follows a long tailed distribution. All posts and comments tagged as public are viewable to all persons within the enterprise. There are multiple public communities on the platform that anchor the topics of discussion in that theme. There are communities such as technology focused, work life balance, diversity and inclusion, specific technologies, specific customers, hobbie,s and many others. Users engage with the platform by posting blogs, microblogs, commenting, liking, sharing, following persons, tags and subscribing to a community. The platform calculates and associates a patented popularity score to each post or comment on the platform. This popularity score relates to the engagement activity in terms of likes, shares, follows, and other similar attributes. 
Linguistically, it is observed that users use semi-formal language and posts are mostly grammatically well formed. Also, it is noted that posts and comments have significantly more positive sentiment than negative.

Our collection includes a small subset collected between 2012 and 2014 from the work life balance community of blogs and their corresponding comments. It contains 10284 users who have posted 2167 blogs that have received 17692 comments. Our collection also contains microblogs and comments. However, we have ignored the microblogs as they do not add vital statistical insights. We have noticed  that users who blog also microblog; thus analyzing blogs alone provides us sufficient information. 

\subsection{Dialogue Pre-Processing}\label{sec:dialoguePreProcessing}

From our dataset collection, we sample a very small section of 10 blogs that have received 229 comments where 127 users have participated. These sample blogs are numerically balanced classes of those that end in a positive sentiment comment (5 blogs) and negative sentiment comment (5 blogs). Within these samples, 3 blogs contain 3 comments with the word `thank' in them; and 3 blogs contain 3 comments with the word `sorry' in them. We extract relevant features from these conversations and use them as inputs to the mining algorithm as discussed in Section \ref{sec:algo}. We have comma-separated-value trace files for each dialogue, i.e., for each blog, for each post and comment in the blog, we have a timestamp, the overall sentiment intensity and polarity of the message (used TextBlob \cite{textblob}). Note that the polarity values range from -5 to 5 with an average value 0.1, the intensity values of each of the 200 EMPATH categories range from 0 to 2 with average values of 1, the number of `thank' words used that range from 0 to 3 with an average of 0, number of `sorry' words used that range from 0 to 2 with an average of 0. We have also added an additional label as to whether the conversation ended on a positive or negative note; this label is based on the sentiment intensity and polarity of the last comment. 


\subsection{Inputs/Parameters for Causal Learning}

As explained in Section~\ref{sec:causal-learning}, the causal inference engine {\tt PSI-Miner} requires learning meta-parameters to decide the explanation template and control over-fitting. In our experiments we use parameters $n=3$, $k=10000$, and a maximum tree depth of 10. The explanation template generated is of the following form:

\begin{center}
{\tt\small$\sqcup_3$ \#\#[0:10$^4$]  $\sqcup_{2}$ \#\#[0:10$^4$] $\sqcup_1$ \#\#[0:10$^4$] $\sqcup_0$|->E }
\end{center}

\begin{table}[t!]
\centering
{\small
\begin{tabular}{ccc}
\toprule
     
     \multirow{2}{*}{\textbf{Attribute}} & \multirow{2}{*}{\textbf{Predicate}} & {\textbf{Value}}\\
      & & \textbf{Domain}\\\midrule
     
     \multirow{4}{*}{sentiment} & {\tt SENTIMENT\_VERY\_NEG} & $<-0.2$\\
     & {\tt SENTIMENT\_NEG} & $[-0.2,0)$\\
     & {\tt SENTIMENT\_LOW\_POS} & $[0,0.5)$\\
     & {\tt SENTIMENT\_MEDIUM\_POS} & $[0.5,0.8)$\\
     & {\tt SENTIMENT\_HIGH\_POS} & $>0.8$ \\
    \midrule
    \multirow{4}{*}{{\tt EMPATH}} & {\tt CATEGORY\_ABSENT} & $=0$\\
    & {\tt CATEGORY\_LOW} & $(0:1)$\\
    & {\tt CATEGORY\_MEDIUM} & $[1:2)$\\
    & {\tt CATEGORY\_HIGH} & $\geq 2$\\
    \midrule
    \multirow{2}{*}{outcome} & {\tt SENTIMENT\_LOW } & $==0$\\
    & {\tt SENTIMENT\_HIGH } & $==1$\\
\bottomrule
\end{tabular}
}
\caption{Predicates used for Causal Learning of Dialogue Outcomes. Predicates are developed over three types of psycho-linguistic attributes, including \textit{sentiment}, \textit{empath categories}, and an \textit{ending sentiment}. For each EMPATH, category four predicates of the form {\tt CATEGORY\_$\langle$LEVEL$\rangle$} are created. The categories include: office, domestic work, medical emergency, aggression, and family. The category levels are {\tt ABSENT}, {\tt LOW}, {\tt MEDIUM} and {\tt HIGH}. Each predicate asserts that the attribute is within a specified value domain (third column).}\label{table:predicates}
\end{table}

\begin{table*}[ht!]
\centering
{\small
\begin{tabular}{cccc}
\toprule
\textbf{Sr.No} & \textbf{Rule} 
& \textbf{Support} & \textbf{Correlation} \\
\midrule
    1 & {\tt \footnotesize !DOMESTIC\_WORK\_ABSENT \&\& MEDICAL\_EMERGENCY\_ABSENT} 
    & 31\% & 55.63\% \\
    \midrule
    
    \multirow{2}{*}{2} & {\tt \footnotesize !FAMILY\_ABSENT \&\& !SENTIMENT\_POS \#\#[0:2$\times 10^4$] !OFFICE\_ABSENT} 
    & 17.39\% & 31.21\%\\
    & {\tt \footnotesize \#\#[0:$10^4$] DOMESTIC\_WORK\_ABSENT \&\& MEDICAL\_EMERGENCY\_ABSENT} & \\
    \midrule
    
    \multirow{3}{*}{3} & {\tt \footnotesize !FAMILY\_ABSENT \&\& !SENTIMENT\_POS \&\& OFFICE\_ABSENT} 
    & \multirow{3}{*}{14.72\%} & \multirow{3}{*}{26.41\%}\\
    & {\tt \footnotesize\#\#[0:2$\times 10^4$] OFFICE\_ABSENT \&\& !FAMILY\_HIGH \&\& !AGGRESSION\_ABSENT}\\
    & {\tt \footnotesize\#\#[0:$10^4$] DOMESTIC\_WORK\_ABSENT \&\& MEDICAL\_EMERGENCY\_ABSENT} \\\midrule
    
    \multirow{4}{*}{4} & {\tt \footnotesize !FAMILY\_ABSENT \&\& !SENTIMENT\_POS \&\& OFFICE\_ABSENT} 
    & \multirow{4}{*}{14.58\%} & \multirow{4}{*}{26.16\%}\\
    & {\tt \footnotesize \#\#[0:3$\times 10^4$] !OFFICE\_ABSENT }\\
    & {\tt \footnotesize \#\#[0:2$\times 10^4$] OFFICE\_ABSENT \&\& !FAMILY\_HIGH \&\& AGGRESSION\_ABSENT}\\
    & {\tt \footnotesize \#\#[0:$10^4$] DOMESTIC\_WORK\_ABSENT \&\& MEDICAL\_EMERGENCY\_ABSENT}\\
    
\bottomrule
\end{tabular}
}
\caption{Explanation Rules Learned for Dialogues with Positive Outcomes in ESM}\label{table:positiveOutcomes}
\end{table*}

\begin{table*}[ht!]
\centering
{\small
\begin{tabular}{cccc}
\toprule
\textbf{Sr.No} & \textbf{Rule} 
& \textbf{Support} & \textbf{Correlation} \\
\midrule
    \multirow{3}{*}{1} & {\tt \footnotesize !FAMILY\_ABSENT \&\& !SENTIMENT\_POS} 
    & \multirow{3}{*}{21.42\%} & \multirow{3}{*}{48.38\%}\\
    & {\tt \footnotesize \#\#[0:3$\times 10^4$] FAMILY\_HIGH \&\& SENTIMENT\_VERY\_POS } \\
    & {\tt \footnotesize \#\#[0:$10^4$] DOMESTIC\_WORK\_ABSENT}\\\midrule
    
    
    
    \multirow{4}{*}{2} & {\tt \footnotesize !FAMILY\_ABSENT \&\& !SENTIMENT\_POS \&\& OFFICE\_ABSENT} 
    & \multirow{4}{*}{21.38\%} & \multirow{4}{*}{48.30\%}\\
    & {\tt \footnotesize \#\#[0:3$\times 10^4$] DOMESTIC\_WORK\_ABSENT \&\& SENTIMENT\_VERY\_POS  }\\
    & {\tt \footnotesize \#\#[0:2$\times 10^4$] !FAMILY\_HIGH }\\
    & {\tt \footnotesize \#\#[0:$ 10^4$] DOMESTIC\_WORK\_ABSENT}\\
    
    
\bottomrule
\end{tabular}
}
\caption{Explanation Rules Learned for Dialogues with Negative Outcomes in ESM}\label{table:negativeOutcomes}
\end{table*}

An explanation can be developed over at most 3$\times 10^4$ time units.  Table~\ref{table:predicates} lists the predicates provided to the causal inference engine for mining explanations rules. The predicates therein express psychologically relevant attributes. 

As our dataset \ref{sec:dataset} is from the community discussing work life balance issues in an enterprise, for our experiments we choose a few features relevant to the discussions around work life balance. Apart from the overall sentiment of the post or comment, we consider the EMPATH categories, \emph{office} (where the author is thinking about their workplace), \emph{domestic work} (where the author is concerned with chores at their home), \emph{medical emergency} (where the author is concerned with a medical emergency), \emph{family} (where the author is thinking about their family), and \emph{aggression} (where the author is demonstrating aggression through their words).

The values for these features are continuous and real-valued. In order to learn interpretable rules, we transform these values and we derive linguistic representations of each feature variable. In particular, we discretize these features to transform them into human interpretable symbolic quantities. It is common in literature for these features to be quantized by equally dividing them based on their observed range. In our experiments however, we use our understanding of the domain.


\subsection{Results}\label{sec:results}

The results of applying causal learning to identify which dialogue events lead to different outcomes in dialogues are summarized in Tables~\ref{table:positiveOutcomes} and~\ref{table:negativeOutcomes}. The tables contain a selection of explanation rules learned. The rules are expressed as temporal logic statements in the form described in Section~\ref{sec:causal-learning}. The temporal intervals are determined by the discretization factor $k$ chosen as a meta-parameter of {\tt PSIMiner}. Note that the value of zero as the left bound of the interval indicates that the dialogue events following the interval can happen either at the same time as the preceding ones or at any time up to the right bound indicated in the interval.

From Table~\ref{table:positiveOutcomes}, we see that the inferred rules are sound from a common-sense perspective. For example, the first rule can be read as "\textit{when at least someone is thinking about their home and no one is worried about any medical emergencies, the dialogue outcome is positive}"; while the second rule can be read as "\textit{when one party commented about their home followed by one party 20000 seconds later talking about the workplace followed by one party not thinking about house work and no one had a medical emergency in the next 10000 seconds results in a positive outcome to the dialogue}". 

From Table~\ref{table:negativeOutcomes}, we see that the inferred rules are sound as well. For example, the first rule can be read as "\textit{one party was not positive followed by one party 30000 seconds later thinking about family while one party was very positive followed 10000 seconds later when they were not discussing their home chores, meanwhile one party in the dialogue has thought about their family at any point in time leads to a negative outcome}". In common sense terms, this rule deals with the idea that despite positive sentiments in a dialogue, not thinking about home chores while thinking about family may cause conflicts and lead to a negative outcome.

The mined rules are both interesting and insightful temporal rules and in our simple setting, causally explain their respective dialogue outcomes.

\section{Conclusions}\label{sec:concl}
Analyzing multi-party dialogues can be difficult due to their inherent complex dynamics. 
The final sentiment is often causally determined by temporal
patterns of psycholinguistic attributes including intermediate sentiments, key
words, likes, etc. across the time span of the communications. In this
work, we characterize a dialogue as a sequence of several symbolic
linguistic variables extracted from dialogue text. The {\tt PSIMiner}
tool is used to derive explanatory temporal rules for prediction of the outcome sentiment. Our pipeline uses a two step unsupervised approach requiring no human annotation. Dialogue attributes such as sentiments and EMPATH categories are mined using pre-trained models. These are then used, unsupervised, by {\tt PSIMiner} to automatically mine dialogue rules. Experimental results are reported for
dialogues in the enterprise social media of a large IT company. The
explanations are found to be plausible and useful. Additional
attributes involving personality traits of participants may be
included in future analysis.


\bibliographystyle{ACM-Reference-Format}
\bibliography{sud.bib}

\end{document}